\documentclass[runningheads]{llncs}

\usepackage{eccv}

\usepackage{eccvabbrv}

\usepackage{graphicx}
\usepackage{booktabs}
\usepackage{pifont}
\usepackage{amssymb}
\usepackage{xcolor}
\usepackage[dvipsnames]{xcolor}

\newcommand{\gray}[1]{\textcolor{gray}{#1}}
\newcommand{\green}[1]{\textcolor{green}{#1}}
\newcommand{\red}[1]{\textcolor{red}{#1}}
\newcommand{\darkred}[1]{\textcolor{red}{#1}}
\newcommand{\cmark}{\ding{51}}%
\newcommand{\xmark}{\ding{55}}%
\usepackage[accsupp]{axessibility}  

\usepackage[pagebackref,breaklinks,colorlinks]{hyperref}

\usepackage{orcidlink}
\usepackage{makecell}
\usepackage{amssymb}
\usepackage{pifont}
\usepackage{setspace}
\usepackage{multicol,multirow}
\usepackage{xcolor}
\usepackage{colortbl}
\usepackage{soul}
\definecolor{ligntgray}{gray}{0.3}
\setuldepth{Berlin}
\begin{document}

\title{Generalizable Neural Human Renderer} 


\author{Mana Masuda \and Jinhyung Park \and Shun Iwase \and  Rawal Khirodkar \and Kris Kitani}

\authorrunning{M. Masuda et al.}

\institute{Carnegie Mellon University\\
\email{\{manam, jinhyun1, siwase, rkhirodk, kmkitani\}@andrew.cmu.edu\\}
}

\maketitle

\begin{figure*}[tb]
    \centering
    \includegraphics[width=1.0\linewidth]{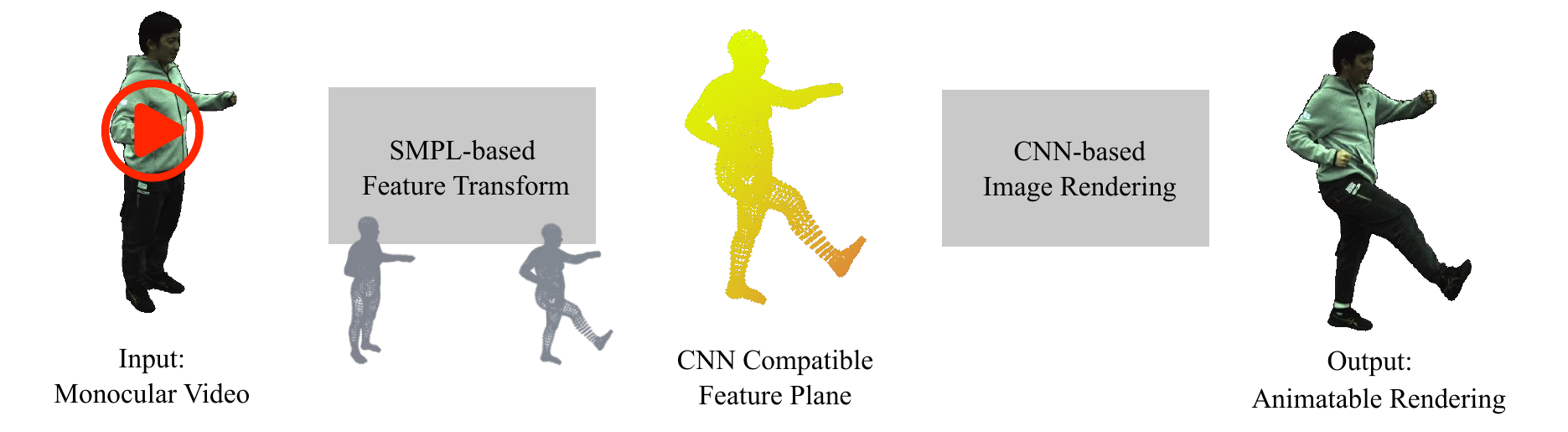}
    \captionsetup{aboveskip=0pt}\captionsetup{belowskip=0pt}
    \caption{
         Given only a monocular video as input, our novel generalizable human rendering framework outputs high-fidelity animatable human rendering without any test-time optimization.
    }
    \label{fig:teaser}
\end{figure*}
\begin{abstract}
  While recent advancements in animatable human rendering have achieved remarkable results, they require test-time optimization for each subject which can be a significant limitation for real-world applications.
  To address this, we tackle the challenging task of learning a Generalizable Neural Human Renderer (\textbf{GNH}), a novel method for rendering animatable humans from monocular video without any test-time optimization.
  Our core method focuses on transferring appearance information from the input video to the output image plane by utilizing explicit body priors and multi-view geometry. To render the subject in the intended pose, we utilize a straightforward CNN-based image renderer, foregoing the more common ray-sampling or rasterizing-based rendering modules.
  Our GNH achieves remarkable generalizable, photorealistic rendering with unseen subjects with a three-stage process.
  We quantitatively and qualitatively demonstrate that GNH significantly surpasses current state-of-the-art methods, notably achieving a $31.3\%$ improvement in LPIPS. 
\end{abstract}
\section{Introduction}
Recent advances in free-viewpoint rendering of humans in motion have achieved remarkable breakthroughs by combining parametric human body models~\cite{loper2015smpl,pavlakos2019expressive,romero2022embodied} and neural radiance fields (NeRF)~\cite{mildenhall2021nerf,barron2021mip,barron2023zip} or 3D Gaussian Splatting \cite{qian20233dgs,hu2023gauhuman}.
Nevertheless, in practical scenarios requiring quick generation, current methods~\cite{liu2021neural,weng2022humannerf} encounter limitations due to the necessity of hours of test-time optimization for each subject. 
Although some recent studies utilize empty space skipping schemes \cite{jiang2023instantavatar} or 3D Gaussian splatting \cite{qian20233dgs} to speed up rendering, the extensive test-time optimization needed for every subtle variation in subjects or clothing severely hinders its wide applicability in dynamic human rendering.

Synthesizing novel views from single-view inputs without the need for subject-specific fine-tuning poses a significant challenge. This is due to the complexities involved in transforming geometry-aware features from the input to the target view, depending on human pose and appearance. Existing generalizable methods~\cite{kwon2022neural,pan2023transhuman,li2023ghunerf,Hu_2023_ICCV} address this challenge by either reconstructing the complete human body in 3D from multi-view inputs~\cite{kwon2022neural,pan2023transhuman} or by inferring missing information from a single input image~\cite{Hu_2023_ICCV}. GHuNeRF~\cite{li2023ghunerf} can reconstruct the human body from a monocular video, but it still suffers from severe blurriness and takes several seconds for rendering, because it adheres to NeRF's conventional ray-sampling rendering process.

To enable high-fidelity animatable human rendering from monocular video without the need for test-time optimization for each subject, we present the Generalizable Neural Human Renderer (\textbf{GNH}). This achievement is facilitated by a streamlined three-stage process: 1) appearance feature extraction, 2) feature transformation, and 3) multi-frame fusion, followed by CNN-based image rendering. Specifically, GNH processes monocular video with known poses and camera views, extracting generalizable appearance features and projecting them into 3D space utilizing an explicit SMPL body prior. This step allows the transfer of appearance information from the input videos to the output image plane. In the second stage, appearance features are transformed to the output image plane using an SMPL-based transformation. GNH then fuses these features into a CNN-compatible image feature through multi-view geometry, to render the subject with a CNN-based image renderer. These components enable GNH to produce high-fidelity images of moving subjects, circumventing the extensive test-time optimization characteristic of many previous methods and achieving faster rendering speeds. 

To assess the performance of GNH, we conduct experiments on three widely-used human datasets: the ZJU-MoCap dataset~\cite{peng2021neural}, the People Snapshot dataset \cite{alldieck2018video}, and the AIST++ dancing dataset~\cite{Li2021aist}. These datasets include multi-view RGB videos of various subjects, showcasing diversity in subject appearance and clothing. The results demonstrate that GNH surpasses contemporary generalizable human rendering methods, improving LPIPS by $31.3\%$ and reducing average rendering error by $17.4\%$ compared to the state-of-the-art. In summary, our contributions are as follows:
\begin{itemize}
    \item We introduce the \textit{Generalizable Neural Human Renderer (GNH)}, an innovative approach designed to render animatable humans from monocular video inputs, eliminating the need for subject-specific test-time optimization.
    \item By transferring appearance features from the input video to the output image plane, GNH yielded superior high-fidelity rendering results compared to previous methods of animatable human rendering.
    \item Comprehensive evaluations reveal that GNH sets a new benchmark in performance, achieving state-of-the-art results. Notably, GNH improves upon previous generalizable human NeRF methods \cite{Hu_2023_ICCV,li2023ghunerf} by reducing LPIPS by $31.3\%$ while also achieving a fourfold increase in rendering speed. Furthermore, GNH surpasses the existing non-generalizable human NeRF approach \cite{weng2022humannerf} by lowering LPIPS by $4.7\%$.
\end{itemize}
\section{Related Works}
\subsection{Novel View Synthesis for Humans}
Free-viewpoint rendering of humans is a challenging task that has been investigated over the last couple of decades \cite{Kanade1997,matusik2000,carranza2003,starck2007,flagg2009,casas2014}. One active stream for rendering humans in novel poses is human pose retargeting \cite{ma2017pose,balakrishnan2018synthesizing,siarohin2018deformable,neverova2018dense,grigorev2019coordinate,chan2019everybody,sarkar2021style,wang2021dance} by using 2D human pose parameters \cite{cao2017realtime,guler2018densepose}.
Liu \etal \cite{liu2019neural} and following works \cite{sarkar2020neural,huang2020arch,he2020geo,he2021arch++,xiu2022icon,alldieck2022photorealistic,iqbal2023rana} start using a pre-captured 3D body model.
Recent advancements in the field of novel view synthesis have been marked by the introduction of the Neural Radiance Field (NeRF) \cite{mildenhall2021nerf} and its follow-up works \cite{barron2021mip,pumarola2021d,park2021nerfies,barron2022mip}.
Following its developments, many NeRF-based animatable human rendering methods have been proposed for both body part rendering \cite{corona2022lisa,guo2023handnerf,kania2023blendfields,chen2023hand,buhler2023preface,trevithick2023real} and full body rendering \cite{peng2021neural,peng2021animatable,xu2021h,noguchi2021neural,kwon2021neural,liu2021neural,su2021nerf,zhao2022humannerf,wang2004image,jiang2022neuman,wang2022arah,su2022danbo,mihajlovic2022keypointnerf,jiang2023instantavatar,shen2023x,sun2023neural,xiang2023rendering,choi2022mononhr,yu2023monohuman}.
Peng \etal \cite{peng2021neural} used learned structured latent codes embodied for reposed mesh vertices from the SMPL model \cite{loper2015smpl} and introduced a NeRF-based neural renderer.
Weng \etal \cite{weng2022humannerf} is one of the first works that takes only monocular video as input. They achieved high-fidelity full 3D rendering by mapping multiple frames of a moving person into a canonical, volumetric T-pose.
InstantAvater\cite{jiang2023instantavatar} achieved in a minute of training and fast rendering speed from an empty space skipping scheme designed for dynamic scenes with known articulation patterns.
Recent advancements in 3D Gaussian Splitting achieved fast training and rendering of animatable head avatars \cite{zhao2024psavatar,saito2023relightable,xu2023gaussian,dhamo2023headgas,chen2023monogaussianavatar,wang2023gaussianhead,xiang2023flashavatar,lan2023gaussian3diff}, hand avatar \cite{jiang20233d} and full-body avatars from multi-view inputs \cite{zielonka2023drivable,moreau2023human,li2023animatable,pang2023ash} or even a monocular video input \cite{kocabas2023hugs,hu2023gaussianavatar,lei2023gart,li2023human101,jena2023splatarmor,qian20233dgs,jung2023deformable,hu2023gauhuman}.

\subsection{Generalizable Novel View Synthesis}
As a downstream task of NeRF, many scene-agnostic (generalizable) novel view synthesis methods have been proposed \cite{yu2021pixelnerf,chen2021mvsnerf,wang2021ibrnet,johari2022geonerf,liu2022neural,varma2023gnt,long2022sparseneus,ren2023volrecon,tian2023mononerf,zhao2023generalized,xu2023wavenerf,Ye_2023_ICCV}.
PixelNeRF \cite{yu2021pixelnerf} employs an image encoder to condition NeRF on 2D image features.
FeatureNeRF \cite{Ye_2023_ICCV} solved this problem by distilling vision foundation models \cite{caron2021emerging,rombach2022high}.
Zhao \etal~\cite{zhao2023generalized} firstly worked on a generalizable method for general dynamic scenes from a monocular video, though this approach assumes small motions and is incapable of expressing dynamic human motions.

\subsection{Generalizable Novel View Synthesis for humans}
Synthesizing novel views of moving humans in a scene-agnostic way is an especially challenging task due to the complexity and diversity of human appearances and poses.
Most of the existing works \cite{raj2021pixel,kwon2021neural,chen2022geometry,gao2022mps,kwon2022neural,gao2023neural,pan2023transhuman} take multiple camera inputs and render novel views of unseen subjects in unseen poses by integrating the features of input camera views by using body model as a geometric prior.
ActorsNeRF \cite{mu2023actorsnerf} trains the network on multiple subjects and synthesizes unseen poses of subjects from monocular video with few-shot optimization on that subject.
Geng \etal \cite{geng2023learning} also presents a generalizable NeRF model to leverage learned prior to reducing the optimization time to minutes.
GHuNeRF \cite{li2023ghunerf} achieved generalizable human rendering from a monocular video by introducing a visibility-aware aggregation scheme.
DINAR \cite{svitov2023dinar} creates full-body avatars from a single image by utilizing a diffusion model to inpainting the unseen part of the body. SHERF \cite{Hu_2023_ICCV} also recovers the full-body avatars from a single image from a combination of three different features.
GPS-Gaussian \cite{zheng2023gps} achieved generalizable and fast full-body rendering from multi-view inputs by utilizing Gaussian-based rendering, but they do not support pose changes.
\section{Method: Generalizable Neural Human Renderer}
\begin{figure*}[tb]
    \centering
    \includegraphics[width=1.00\linewidth]{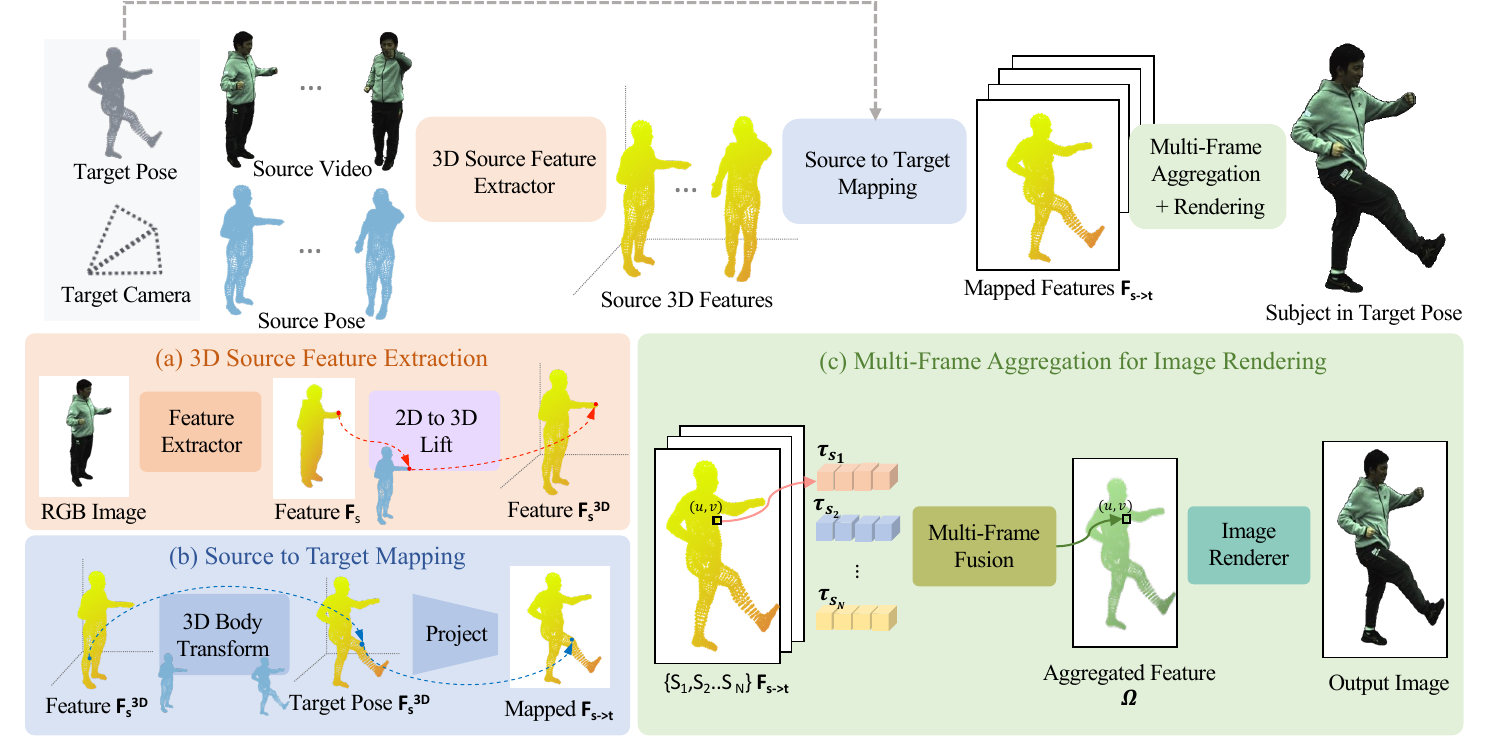}
    \captionsetup{aboveskip=3pt}\captionsetup{belowskip=0pt}
    \caption{
        Overview of our Generalizable Neural Human Renderer (GNH).
        \textit{Stage 1 - 3D Source feature extraction} obtains 3D source feature using 2D feature extraction and lifting it to 3D using the body mesh vertices and camera parameters. \textit{Stage 2 - Source to target mapping} converts the 3D source features to the target domain and projects them into 2D.
        \textit{Stage 3 - Multi-frame aggregation and rendering} consolidates information from all the source frames using multi-view geometry and renders the target image using a CNN-based renderer.
    }
    \label{fig:overview}
\end{figure*}

We propose the Generalizable Neural Human Rnederer (GNH) to achieve animatable human rendering solely from a monocular video of the subject. The overall pipeline of our GNH is shown in Fig. \ref{fig:overview}.

\textbf{Problem Setting}
Given a monocular video, estimated camera pose, subject pose, target camera pose, and target subject pose, we first sample a set of $N$ frames as source input ${\{\mathbf{I}_s \in \mathbb{R}^{H \times W \times 3}, \mathbf{P}_s \in \mathbb{R}^{3 \times 4}, \boldsymbol{\theta}_s \in \mathbb{R}^{24 \times 3}\}}_{s=1}^N$, where $\mathbf{I}_s$ represents the source image, $\mathbf{P}_s$ the camera extrinsic, and $\boldsymbol{\theta}_s$ the SMPL body pose parameters \cite{loper2015smpl}. Given this input, we render the subject in a specified pose $\boldsymbol{\theta}_t$ from a queried target camera pose $\mathbf{P}_t$.
\subsection{Source Feature Extraction}
\hspace{\parindent}\textbf{2D Source Image Feature Extraction.}
As seen in Fig.\ref{fig:overview}(a), the architecture of our source feature extraction begins with a convolutional feature extraction backbone. We leverage a pre-trained DINO-v2 \cite{oquab2023dinov2} as a robust extractor for low-resolution image features, $\boldsymbol{\phi}_\text{coarse}$. To better align this feature to our task, we then refine and upsample this image feature by feeding it into a vision transformer $\boldsymbol{\phi}_\text{refine}$, which produces a refined coarse image feature \cite{xie2021segformer,trevithick2023real}:
\begin{equation}
    \mathbf{F}_\text{coarse} = \boldsymbol{\phi}_\text{refine}(\boldsymbol{\phi}_\text{coarse}(\mathbf{I}_\text{s})).
\end{equation}
This coarse feature extraction stream has a parallel fine-grained feature extraction stream, $\boldsymbol{\phi}_\text{fine}$, using convolutional layers with a single downsampling step:
\begin{equation}
    \mathbf{F}_\text{fine} = \boldsymbol{\phi}_\text{fine}(\mathbf{I}_\text{s}),
\end{equation}
which aims to maintain a high-resolution feature map to preserve high-fidelity details for the target image rendering. 
The final per-source frame feature $\mathbf{F}_\text{s}$ is generated by channel-wise concatenation of the coarse image feature and fine image feature:
\begin{equation}
    \mathbf{F}_\text{s} = \mathbf{F}_\text{coarse} \oplus \mathbf{F}_\text{fine}.
\end{equation}

\textbf{2D-to-3D Feature Lifting.}
Following the 2D feature extraction module, our source feature extraction module lifts this 2D feature into 3D space. To handle large human motion while keeping generalizability, we leverage the SMPL body template to seamlessly transfer the source frame feature into the 3D space. We extract a 3D mesh $\mathcal{M}_s$ from the source human pose parameters $\boldsymbol{\theta}_s$, containing mesh vertices $\{\boldsymbol{v}_s^{i}\}_{i=1}^{6890}$. Note that we set SMPL body shape parameters as $\boldsymbol{\beta} = \boldsymbol{0}$ to keep generalizability for various human shapes. Each mesh vertex is then projected onto the source image feature plane using the source camera's projection function, $\Pi_s$. This projection, $\Pi_s(\boldsymbol{v}_s^{i})$, enables the extraction of an appearance feature $\boldsymbol{F}^{3D}_s(\boldsymbol{v}_s^i)$ from the estimated source image feature $\mathbf{F}_\text{s}$. Thus, each 3D source vertex $\boldsymbol{v}_s^{i}$ is associated with the following appearance feature:
\begin{equation}
    \boldsymbol{F}^{3D}_s(\boldsymbol{v}_s^i) = \mathbf{F}_{s}\big(\Pi_s(\boldsymbol{v}_s^{i})\big).
\end{equation}

\subsection{Source-to-Target Feature Mapping}
\hspace{\parindent}\textbf{Mapping to Target Space.}
As shown in Fig. \ref{fig:overview}(b), our second step consists of transforming the 3D source appearance feature $\boldsymbol{F}^{3D}_s$ into the target pose. We transform the source SMPL mesh vertices to the target space using the given target subject pose parameters. After preparing target 3D mesh $\mathcal{M}_t$ from the given target human pose parameters $\boldsymbol{\theta}_t$ and template body shape parameter $\boldsymbol{\beta} = \boldsymbol{0}$, each target mesh vertex $\boldsymbol{v}_t^{i}$ carries a latent feature vector derived from the source feature $\boldsymbol{F}^{3D}_s(\boldsymbol{v}_s^i)$.

\textbf{Target Feature Projection.}
After transforming each source appearance feature related to $N$ source frame images, the 3D appearance features populate their own 2D target features using the projection function $\Pi_t(\boldsymbol{v}_t^{i})$ of the target camera parameter. At this stage, only half of the SMPL vertices are visible from the target camera view. To account for occlusions in the target features, we only project the vertices visible from the target camera view to the 2D target image plane. The 2D target feature map is formalized as 
\begin{equation}
    \mathbf{F}_{\text{s} \rightarrow \text{t}}\big(\Pi_t(\boldsymbol{v}_t^{i})\big) = \boldsymbol{F}^{3D}_s(\boldsymbol{v}_t^i).
\end{equation}
After this stage, we obtain a target feature map for each source frame, setting the stage for the subsequent rendering process.

\subsection{Multi-Frame Aggregation and Rendering}
The third and final stage of GNH focuses on fusing information from all source images in the context of the target pose. 

\textbf{Multi-Frame Fusion.}
As shown in Fig. \ref{fig:overview}(c), our feature fusion transformer $\Psi_\text{multi\_frame}$ fuses the target features $\mathbf{F}_{\text{s}_i \rightarrow \text{t}}$ from each source frame $s_i \in \{s_1 \dots s_N\}$ into a single target feature map which is compatible with CNN-based image rendering.
Intuitively, this transformer highlights features from source images relevant to the target pose and attenuates features from distant source frames. 
For each pixel $(u,v)$ in the target image, $\Psi_\text{multi\_frame}$ accepts $N$ tokens as input---one from each source frame. Each token corresponds to the latent feature $\mathbf{F}_{\text{s}_i \rightarrow \text{t}}(u, v)$ at that specific pixel location $(u, v)$. This is formally represented as follows:
\begin{align}
\boldsymbol{\tau}_{s_i} &= \mathbf{F}_{\text{s}_i \rightarrow \text{t}}(u, v), \\
\boldsymbol{\Omega}_{u,v} &= \Psi_\text{multi\_frame}(\boldsymbol{\tau}_{s_1}, \boldsymbol{\tau}_{s_2}, \dots, \boldsymbol{\tau}_{s_N}),
\end{align}
where $\boldsymbol{\tau}_{s_i}$ is the $i^\text{th}$ input token to $\Psi_\text{multi\_frame}$, indicating the feature from source $s_i$ mapped to the pixel location in target $t$. $\boldsymbol{\Omega}_{u,v}$ signifies the fused multi-frame feature at each pixel. 

\textbf{Image Rendering.}
The multi-frame fusion transformer outputs a single CNN-compatible feature map $\boldsymbol{\Omega}$ for the target pose, having adaptively aggregated features from each source frame. GNH processes this feature map with a CNN-based image rendering network to generate the target output image. Specifically, the image rendering network within GNH adopts a deep residual U-Net architecture, denoted as $\mathcal{R}$:
\begin{align}
    \hat{\mathbf{I}}_t &= \mathcal{R}(\boldsymbol{\Omega}).
\end{align}

\subsection{Optimizing a Generalizable Neural Human Renderer}
The learnable modules within our proposed GNH framework include the feature refinement module $\boldsymbol{\phi}_\text{refine}$, the fine image feature extractor $\boldsymbol{\phi}_\text{fine}$, the multi-frame fusion transformer $\Psi_\text{multi\_frame}$, and the image renderer $\mathcal{R}$. 
During training, we randomly sample the pairs of $N$ source frames and a target frame set for each training subject. By inputting the source frames into the above-described three-stage process, we aim to reconstruct the subject in the target pose. Four loss functions are used to supervise the training.

\textbf{Photometric Loss.}
Given the ground truth target image $\mathbf{I}_t$ and predicted image $\hat{\mathbf{I}_t}$, we apply the photometric loss as follows:
\begin{equation}
    \mathcal{L}_\text{color} = ||\mathbf{I}_t - \hat{\mathbf{I}}_t||_1^1.
\end{equation}

\textbf{Perceptual Loss.}
We also leverage the perceptual loss LPIPS to ensure the quality of the rendered image:
\begin{equation}
    \mathcal{L}_\text{LPIPS} = \text{LPIPS}(\mathbf{I}_t, \hat{\mathbf{I}}_t).
\end{equation}

\textbf{Adversarial Loss.}
We employed adversarial training for our training procedure. This adversarial training contributes to rendering detailed parts such as clothing patterns and facial features, helping to the enhancement of image resolution:
\begin{equation}
    \mathcal{L}_\text{adv} = \log (1-\mathcal{D}(\hat{\mathbf{I}}_t)),
\end{equation}
where $\mathcal{D}$ denotes our discriminator module.

\textbf{Anti-Bias Loss.}
We also employed an anti-bias loss. This anti-bias loss effectively eliminates color shifts and is robust to minor misalignments. Although perceptual and adversarial losses improve high-frequency details, they may introduce color shifts \cite{prinzler2023diner}:
\begin{equation}
    \mathcal{L}_\text{ab} = ||\text{DS}_k(\hat{I}_t) - \text{DS}_k(I_t)||_1^1, 
\end{equation}
where $\text{DS}_k(\cdot)$ denotes $k$-fold downsampling. 

In summary, the overall loss function is as follows:
\begin{equation*}
    \mathcal{L} = \lambda_1 \mathcal{L}_\text{color} + \lambda_2 \mathcal{L}_\text{LPIPS} + \lambda_3 \mathcal{L}_\text{adv} + \lambda_4 \mathcal{L}_\text{ab},
\end{equation*}
where $\lambda_i$ denotes the weights of each loss.
\section{Experiments}

\subsection{Experimental Setup}

\hspace{\parindent} \textbf{Datasets.}
We evaluated GNH using three widely used datasets: the ZJU-MoCap~\cite{peng2021neural}, the People Snapshot~\cite{alldieck2018video}, and the AIST++ dance motion~\cite{Li2021aist}. For ZJU-MoCap, we used six subjects recorded with ``camera 1'' for training, leaving three subjects (387, 393, 394) for evaluation.
In the People Snapshot dataset, 21 videos were utilized for training with three videos (female-3-casual, male-1-casual, male-3-casual) set aside for evaluation.
The AIST++ dataset, featuring nine multi-view videos of 30 subjects in dynamic dance scenes, was used by selecting one action sequence per subject for training with 25 subjects and evaluating on five subjects (16-20). 
We employed PointRend \cite{kirillov2020pointrend} to generate foreground masks for AIST++.
The test set involved sampling the initial part of each video for source input frames and the remainder for evaluation in the task of animatable human rendering. For further details, please refer to the supplemental materials.

\textbf{Baselines.}
We compare our method against state-of-the-art methods for modeling humans in motion, including HumanNeRF \cite{weng2022humannerf} (test-time optimization required, single video), ActorsNeRF \cite{mu2023actorsnerf} (generalizable training and test-time optimization required, single video), Neural Human Performer (NHP) \cite{kwon2022neural} (multi-view video required), SHERF \cite{Hu_2023_ICCV} (fully generalizable method, single image input), and GHuNeRF \cite{li2023ghunerf} (fully generalizable method, single video input).

\textbf{Evaluation Metrics.}
To quantitatively evaluate the quality of the rendered images, we report on three widely-used metrics: Peak Signal-to-Noise Ratio (PSNR), Structural Similarity Index (SSIM) \cite{wang2004image}, and Learned Perceptual Image Patch Similarity (LPIPS) \cite{zhang2018unreasonable}.
To facilitate easier comparison in certain evaluation setups, we set an ``average'' error metric \cite{barron2021mip} that summarizes all three metrics: the geometric mean of three metrics: Mean Squared Error $\mathrm{MSE}=10^{-\mathrm{PSNR}/10}$, dissimilarity measure $\mathrm{dssim} = \sqrt{1-\mathrm{SSIM}}$ \cite{Lombardi2019}, and LPIPS.
We report LPIPS and ``average'' errors as multiplied by $\times 10^3$.

\textbf{Implementation Details.}
We use the ResUnet++ \cite{jha2019resunetplus} with group normalization layers \cite{wu2018group} for our image rendering network $\mathcal{R}$. For our adversarial training, we employed the discriminator from Pix2Pix \cite{isola2017image}. For the coarse image feature, we utilize DINO-v2 \cite{oquab2023dinov2} with ViT-S/14 architecture and set the patch size as 64.

\textbf{Optimization Details.}
Our framework contains five different modules to optimize: the coarse feature refinement module, the fine image feature extractor, the multi-frame feature fusion transformer, the image rendering module, and the discriminative module for our adversarial training.
We adopt the Adam optimizer \cite{kingma2014adam} for training with a learning rate of $1e^{-3}$ for the feature encoder module, $5e^{-4}$ for the feature fusion module and the image rendering module, and $1e^{-5}$ for the discriminator module with a batch size of five (other Adam hyperparameters are left at default values of $\beta_1 = 0.9$, $\beta_2 = 0.999$, and $\epsilon=10^{-7}$). We train our network for 4k epochs for the ZJU-MoCap dataset, 7k epochs for the People Snapshot dataset, and 6k epochs for the AIST++ dataset on one RTX 3090 GPU (about 6-7 hours). For all datasets, we set $\{\lambda_1, \lambda_2, \lambda_3, \lambda_4\} = \{0.2, 0.1, 0.05, 0.8\}$ for our objective function. We used $k=2$ for our anti-bias loss.
For the training set, we sampled one frame from the training video as the target and randomly sampled source frames from the rest of the frames. We set $N=5$ during the training for all datasets. 

\textbf{Evaluation Details.}
During the evaluation, we sampled the source frames from the first part of the evaluation video. To use the best feature for the target pose, we sampled the most closely posed frame for the source frames by calculating the torso rotation difference and Procrustes pose difference.
We also set $N=5$ for all evaluation setups except for Table. \ref{tab:n_source_views} and Fig. \ref{fig:n_source_views}.
Following previous works, the evaluations are conducted with an image size of $512 \times 512$ for the ZJU-MoCap dataset and People Snapshot dataset, $540 \times 960$ for the AIST++ dataset.

\begin{table}[tb]
    \centering
    \captionsetup{aboveskip=0pt}\captionsetup{belowskip=0pt}\caption{Quantitative comparison on the ZJU-MoCap dataset \cite{peng2021neural}. (*indicates the numbers are from the original paper.)}
    \resizebox{1.0\linewidth}{!}{
    \begin{tabular}{l|cc|cccc|cccc|cccc}
    \Xhline{3\arrayrulewidth}
      & Multi & Test-time  & \multicolumn{4}{c|}{Person387} & \multicolumn{4}{c|}{Person393} & \multicolumn{4}{c}{Person394} \\ \cline{4-15}
       & View & Opt. & PSNR$\uparrow$ & SSIM$\uparrow$ & LPIPS$\downarrow$ & Avg. $\downarrow$&  PSNR$\uparrow$ & SSIM$\uparrow$ & LPIPS$\downarrow$ & Avg. $\downarrow$ & PSNR$\uparrow$ & SSIM$\uparrow$ & LPIPS$\downarrow$ & Avg. $\downarrow$\\ \hline
       \gray{HumanNeRF} \cite{weng2022humannerf}  & \green{\xmark} & \darkred{\cmark} & \gray{27.25} & \gray{0.960} & \gray{44.68} & \gray{25.83} & \gray{28.05} & \gray{0.960} & \gray{39.97} & \gray{23.38} & \gray{28.51} & \gray{0.962} & \gray{36.02} & \gray{21.65}  \\
       \gray{ActorsNeRF*} \cite{mu2023actorsnerf} & \green{\xmark} & \darkred{\cmark} & \gray{27.61} & \gray{0.961} & \gray{36.18} &   \gray{-}   & \gray{27.59} & \gray{0.957} & \gray{39.36} & \gray{-} & \gray{28.98} & \gray{0.961} & \gray{34.17} & \gray{-}  \\ 
       \gray{NHP} \cite{kwon2022neural} & \darkred{\cmark} & \green{\xmark} & \gray{28.28} & \gray{0.965} & \gray{59.77} & \gray{25.60} & \gray{29.86} & \gray{0.967} & \gray{54.41} & \gray{21.75} & \gray{29.93} & \gray{0.966} & \gray{52.11} & \gray{21.56}\\ \hline \hline
       SHERF \cite{Hu_2023_ICCV} & \green{\xmark} & \green{\xmark} & 25.69 & 0.957 & 56.31 & 31.57 & 27.74 & 0.965 & 39.88 & 23.23 & 28.27 & 0.964 & 42.67 & 23.05\\
       GHuNeRF \cite{li2023ghunerf} & \green{\xmark} & \green{\xmark} & 27.08 & 0.960 & 59.59 & 28.76 &  28.59 & 0.965 & 57.01 & 24.60 &  28.22 & 0.963 & 53.32 & 25.06\\
       GNH (Ours)  & \green{\xmark} & \green{\xmark} & \textbf{27.58} & \textbf{0.963} & \textbf{41.26} & \textbf{23.95} & \textbf{29.15} & \textbf{0.968} & \textbf{39.08} & \textbf{20.52} & \textbf{29.58} & \textbf{0.969} & \textbf{34.22} & \textbf{18.89}
\\
    \hline
    \end{tabular}}
    \label{tab:zjumocap}
\end{table}
\begin{figure*}[tb]
    \centering
    \includegraphics[width=1.0\linewidth]{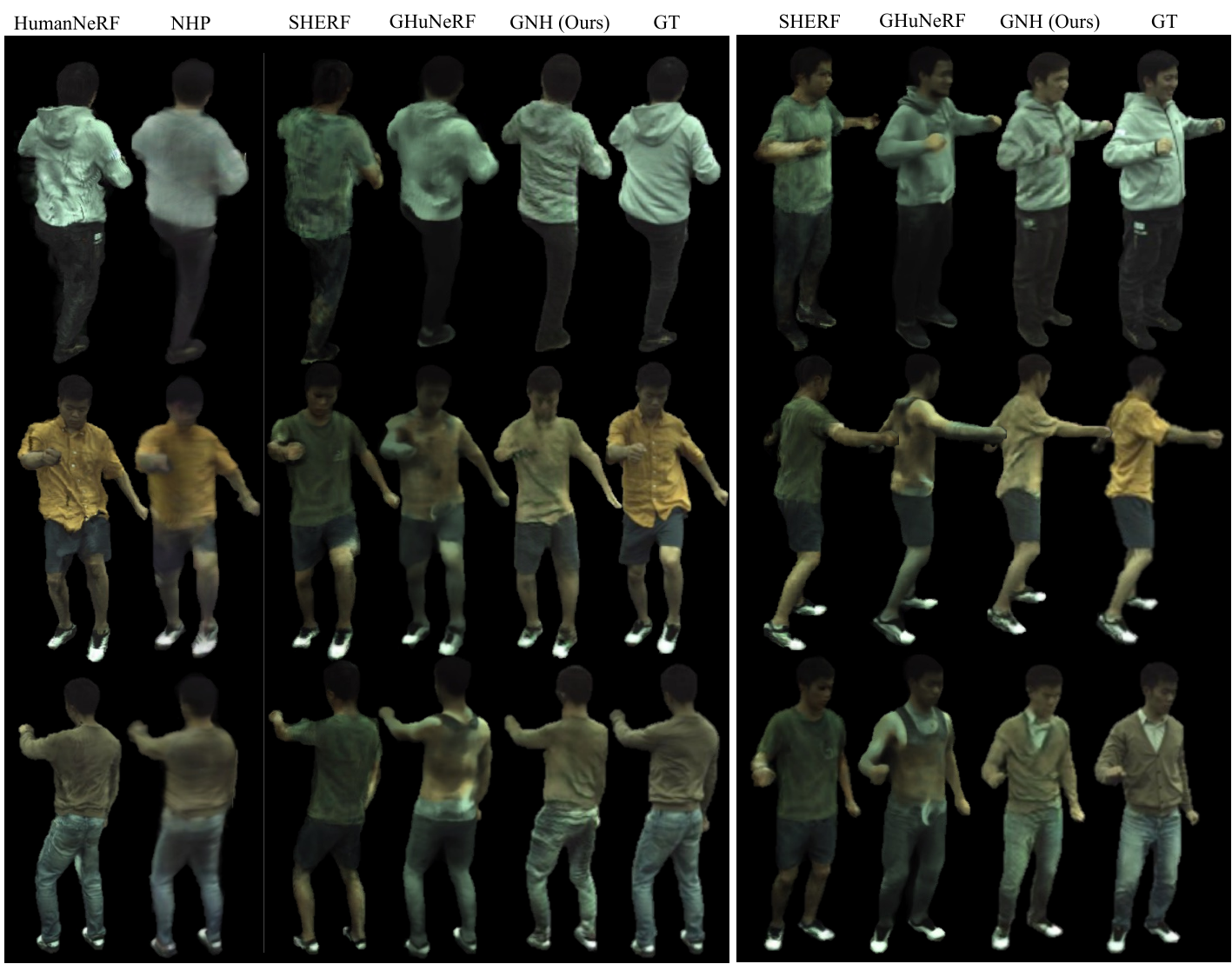}
    \captionsetup{aboveskip=3pt}\captionsetup{belowskip=0pt}
    \caption{
        Qualitative comparison of animatable rendering of \textbf{unseen} identity on the ZJU-MoCap dataset \cite{peng2021neural}.
    }
    \label{fig:zjumocap}
\end{figure*}

\begin{table}[tb]
    \centering
    \captionsetup{aboveskip=0pt}\captionsetup{belowskip=0pt}\caption{Quantitative comparison on the People Snapshot dataset \cite{alldieck2018video}.}
    \resizebox{1.0\linewidth}{!}{
    \begin{tabular}{l|cccc|cccc|cccc}
    \Xhline{3\arrayrulewidth}
      & \multicolumn{4}{c|}{Female-3-Casual} & \multicolumn{4}{c|}{Male-1-Casual} & \multicolumn{4}{c}{Male-3-Casual} \\ \cline{2-13}
       & PSNR$\uparrow$ & SSIM$\uparrow$ & LPIPS$\downarrow$ & Avg. $\downarrow$&  PSNR$\uparrow$ & SSIM$\uparrow$ & LPIPS$\downarrow$ & Avg. $\downarrow$ & PSNR$\uparrow$ & SSIM$\uparrow$ & LPIPS$\downarrow$ & Avg. $\downarrow$\\ \hline
       GHuNeRF \cite{li2023ghunerf} & \textbf{29.63} & 0.962 & 90.97 & 26.89 &  26.15 & 0.939 & 99.77 & 39.10 &  \textbf{27.93} & \textbf{0.952} & 99.47 & 32.09\\
       GNH (Ours)  &  {29.60} & \textbf{0.964} & \textbf{67.41} & \textbf{24.18} & \textbf{27.50} & \textbf{0.944} & \textbf{75.00} & \textbf{31.74} & {27.20} & {0.950} & \textbf{76.06} & \textbf{32.29}
\\
    \hline
    \end{tabular}}
    \label{tab:snapshot}
\end{table}
\begin{figure}[tb]
    \centering
    \includegraphics[width=1.00\linewidth]{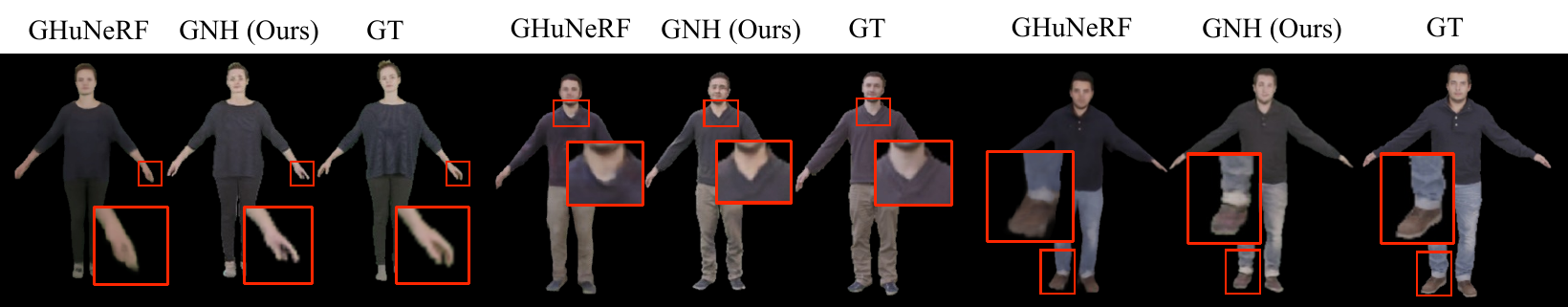}
    \captionsetup{aboveskip=3pt}\captionsetup{belowskip=0pt}
    \caption{
        Qualitative comparison of animatable rendering of \textbf{unseen} identity on the People Snapshot dataset \cite{alldieck2018video}.
    }
    \label{fig:snapshot}
\end{figure}

\begin{table}[tb]
    \centering
    \captionsetup{aboveskip=0pt}\captionsetup{belowskip=3pt}
    \caption{Quantitative comparison on the AIST++ dataset \cite{Li2021aist}. (*indicates the numbers are from the original paper.)}
    \resizebox{1.0\linewidth}{!}{
    \begin{tabular}{l|c|cccc|cccc|cccc}
    \Xhline{3\arrayrulewidth}
      & Test-time & \multicolumn{4}{c|}{Person 16} & \multicolumn{4}{c|}{Person 17} & \multicolumn{4}{c}{Person18} \\ \cline{3-14}
      & Opt. & PSNR$\uparrow$ & SSIM$\uparrow$ & LPIPS$\downarrow$ & Avg. $\downarrow$&  PSNR$\uparrow$ & SSIM$\uparrow$ & LPIPS$\downarrow$ & Avg. $\downarrow$ & PSNR$\uparrow$ & SSIM$\uparrow$ & LPIPS$\downarrow$ & Avg. $\downarrow$ \\ \hline
       \gray{ActorsNeRF*} \cite{mu2023actorsnerf} & \darkred{\checkmark}  & \gray{25.73} & \gray{0.981} & \gray{18.93} &  \gray{ - }  & \gray{26.14} & \gray{0.983} & \gray{18.37} & \gray{ - }
 & \gray{25.03} & \gray{0.983} & \gray{18.52} &  \gray{ - }   \\
       GHuNeRF \cite{li2023ghunerf} & \green{\xmark} & 24.14 & 0.981 & 24.67 & 23.66 & 24.79 & 0.982 & 27.08 & 23.13 & 23.12 & 0.980 & 28.14 & 27.28 \\
       GNH (Ours) & \green{\xmark} & \textbf{27.56} & \textbf{0.987} & \textbf{14.40} & \textbf{14.31} & \textbf{27.84} & \textbf{0.988} & \textbf{15.87} & \textbf{14.27} & \textbf{27.12} & \textbf{0.987} & \textbf{14.89} & \textbf{14.98} 
\\     \hline
   &Test-time & \multicolumn{4}{c|}{Person 19} & \multicolumn{4}{c|}{Person 20} &  & & & \\ \cline{3-10}
      & Opt. & PSNR$\uparrow$ & SSIM$\uparrow$ & LPIPS$\downarrow$ & Avg. $\downarrow$&  PSNR$\uparrow$ & SSIM$\uparrow$ & LPIPS$\downarrow$ & Avg. $\downarrow$ &  & &  & \\ \cline{1-10}
       \gray{ActorsNeRF*}& \darkred{\checkmark} & \gray{25.88} & \gray{0.983} & \gray{18.58} &   \gray{ - }   & \gray{25.78} & \gray{0.985} & \gray{17.44} & \gray{ - } \\ 
       GHuNeRF \cite{li2023ghunerf} & \green{\xmark}  & 24.68 & 0.981 & 23.69 & 22.39 & 26.20 & 0.958 & 22.21 & 18.78  &   & & & \\
       GNH (Ours)  & \green{\xmark}&\textbf{28.44} & \textbf{0.987} & \textbf{13.13} & \textbf{12.90} & \textbf{30.05} & \textbf{0.992} & \textbf{10.61} & \textbf{9.89} &  &  &  &  \\ \cline{1-10}
    \end{tabular}}
    \label{tab:aist}
\end{table}

\begin{figure*}[tb]
    \centering
    \includegraphics[width=1.00\linewidth]{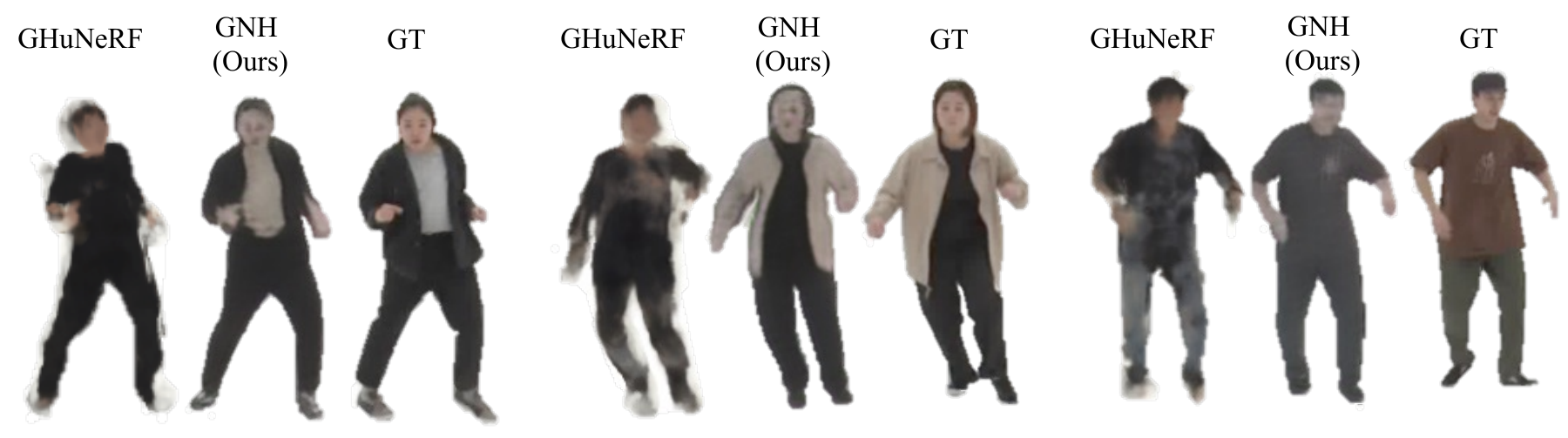}
    \captionsetup{aboveskip=0pt}\captionsetup{belowskip=0pt}
    \caption{
        Qualitative comparison of animatable rendering of \textbf{unseen} identity on the AIST++ dataset \cite{Li2021aist}.
    }
    \label{fig:aist}
\end{figure*}

\vspace{-3mm}
\subsection{Result}
\hspace{\parindent}\textbf{Qualitative Evaluation.}
The qualitative comparisons with baseline methods are shown in Fig. \ref{fig:zjumocap}, Fig. \ref{fig:snapshot}, and Fig. \ref{fig:aist}.
Our GNH demonstrates superior rendering performance both overall and in detail for all subjects compared to SHERF and GHuNeRF. In Fig. \ref{fig:zjumocap}, we observe that SHERF and GHuNeRF tend to overfit the training set and render incorrect individuals during inference. In contrast, GNH produces results that are visually on par with or superior to HumanNeRF~\cite{weng2022humannerf} and NHP~\cite{kwon2022neural}, despite the latter requiring test-time optimization and multi-viewpoint inputs, respectively. 
Fig. \ref{fig:snapshot} demonstrates that our GNH can render a more detailed structure precisely than GHuNeRF. We also found that GHuNeRF cannot render meaningful output in our experiments as shown in Fig. \ref{fig:aist}, likely because the AIST++ dataset is more difficult compared to the other two datasets since it contains dynamic dancing motions.

\textbf{Quantitative Evaluation.}
Our GNH significantly surpasses baseline methods in quantitative evaluations across three key datasets as shown in Table \ref{tab:zjumocap}, Table \ref{tab:snapshot}, and Table \ref{tab:aist}. In LPIPS scores, GNH achieves substantial improvements: $31.5\%$ better than GHuNeRF on ZJU-MoCap, $24.6\%$ on People Snapshot, and $45.2\%$ on AIST++. Against SHERF, GNH shows a $17.5\%$ improvement on ZJU-MoCAP. Remarkably, GNH also outperforms HumanNeRF by $4.7\%$ on ZJU-MoCap, demonstrating its efficiency without necessitating test-time optimization.

\begin{figure*}[tb]
    \centering
    \includegraphics[width=1.00\linewidth]{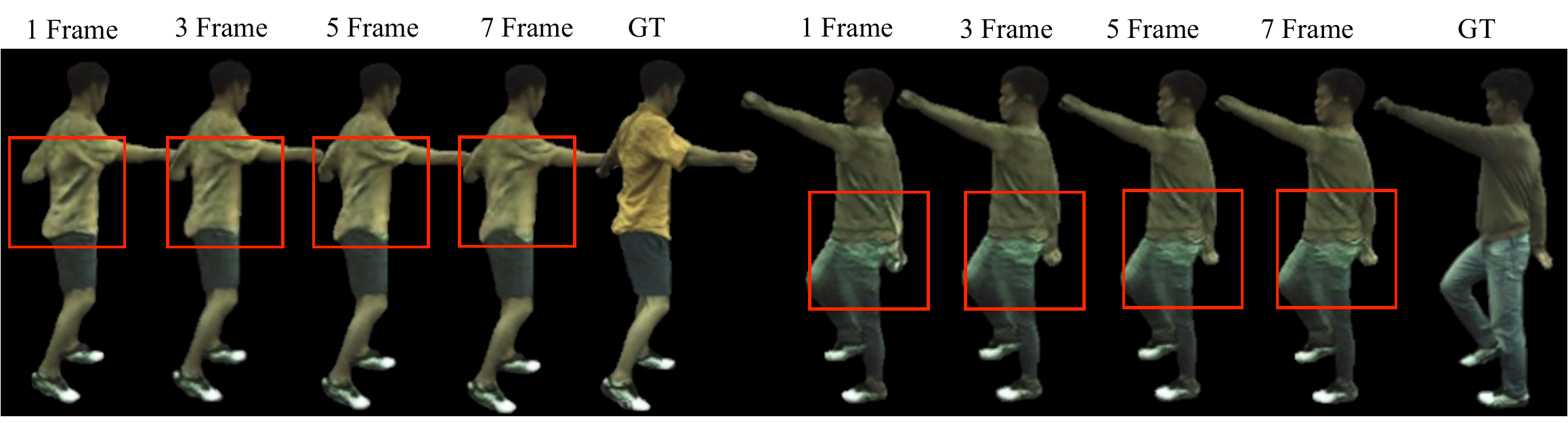}
    \captionsetup{aboveskip=3pt}\captionsetup{belowskip=0pt}
    \caption{
        Qualitative comparison across various numbers of source frames.
    }
    \label{fig:n_source_frames}
\end{figure*}

\textbf{Number of Input Source Frames.}
Our quantitative evaluation on the ZJU-MoCap dataset, considering a range of source frames (1, 3, 5, 7), demonstrates a clear trend in Table \ref{tab:n_source_views}: as the number of input frames increases, the average rendering error notably decreases. This trend, calculated from the averages for three test subjects (People 387, 393, 394), demonstrates our method's efficiency in handling multiple frames. Notably, even with a single frame input, our method outperforms SHERF, a technique that aims to reconstruct unseen parts for animating humans from a single image. A qualitative comparison in Fig. \ref{fig:n_source_frames} further illustrates this advantage, showing our method's superior capability in rendering animatable humans with greater accuracy and detail.

\begin{figure*}[tb]
    \centering
    \includegraphics[width=0.55\linewidth]{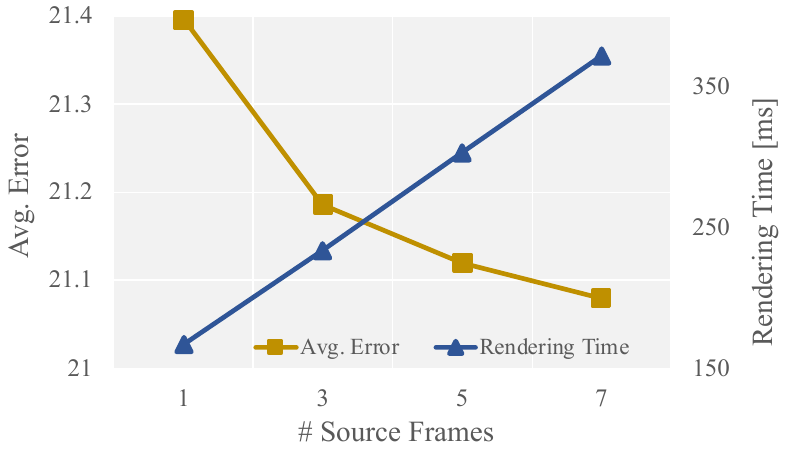}
    \captionsetup{aboveskip=0pt}\captionsetup{belowskip=0pt}
    \caption{
        Rendering speed vs. ``average'' rendering error across various numbers of source frames on the ZJU-MoCap dataset.
    }
    \label{fig:n_source_views}
\end{figure*}

\textbf{Rendering Time.}
A key advantage of generalizable rendering is its ability to complete rendering in a single forward pass, which is crucial for reducing inference time in downstream applications.
We report and compare the inference frames per second (FPS) of GNH across various numbers of source frames with the comparison to baseline generalizable human NeRF methods \cite{Hu_2023_ICCV,li2023ghunerf} in Table \ref{tab:n_source_views}. 
Although the number of input frames and rendering performance are inversely related, GNH achieves a rendering speed that is $2\sim7$ times faster than baseline methods.
Fig. \ref{fig:n_source_views} shows the trade-off relationship between the rendering speed and the ``average'' rendering error. Increasing the number of input frames improves rendering performance, but it results in slower rendering speeds. Conversely, reducing the number of input frames speeds up the rendering process but at the expense of performance.

\begin{table}[tb]
    \centering
    \captionsetup{aboveskip=0pt}\captionsetup{belowskip=0pt}
    \caption{Quantitative evaluation on the ZJU-MoCap dataset across various numbers of source frames. The numbers in parentheses indicate the difference compared to GNH with a 5-frame input result.}
    \scalebox{0.8}{
    \begin{tabular}{l|c|ccccc|cc}
    \Xhline{3\arrayrulewidth}
         Model & \multicolumn{1}{c|}{\#Frames}  & PSNR$\uparrow$ & SSIM$\uparrow$ & LPIPS*$\downarrow$ & \multicolumn{2}{c|}{Avg.* $\downarrow$} & \multicolumn{2}{c}{FPS $\uparrow$}  \\ \hline 
         SHERF \cite{Hu_2023_ICCV} & 1 & 27.23 & 0.962 & 46.29 & 25.95 & \red{(+4.83)} & 1.33 & \red{($\times$0.40)}\\
         GHuNeRF \cite{li2023ghunerf} & 15 & 28.10 & 0.963 & 55.75 & 25.58 & \red{(+4.46)} & 0.76 & \red{($\times$0.23)}\\ \hline \hline
         \multirow{4}{*}{GNH (Ours)}& 1 & 28.66 & 0.966 & 38.19 & 21.39 & \red{(+0.27)} & 5.98 &\green{($\times$1.81)}\\
         & 3 & 28.73 & 0.966 & 38.05 & 21.19 & \red{(+0.07)} & 4.28 & \green{($\times$1.30)}\\
         & 5 & 28.77 & 0.967  & 38.18 & 21.12 & (+0.00) & 3.30 & ($\times$1.00)\\
         & 7 & 28.80 & 0.967 & 38.32 & 21.08 & \green{(-0.04)} & 2.68 & \red{($\times$0.81)}\\ \hline
    \end{tabular}}
    \label{tab:n_source_views}
\end{table}

\begin{figure*}[tb]
    \centering
    \includegraphics[width=1.00\linewidth]{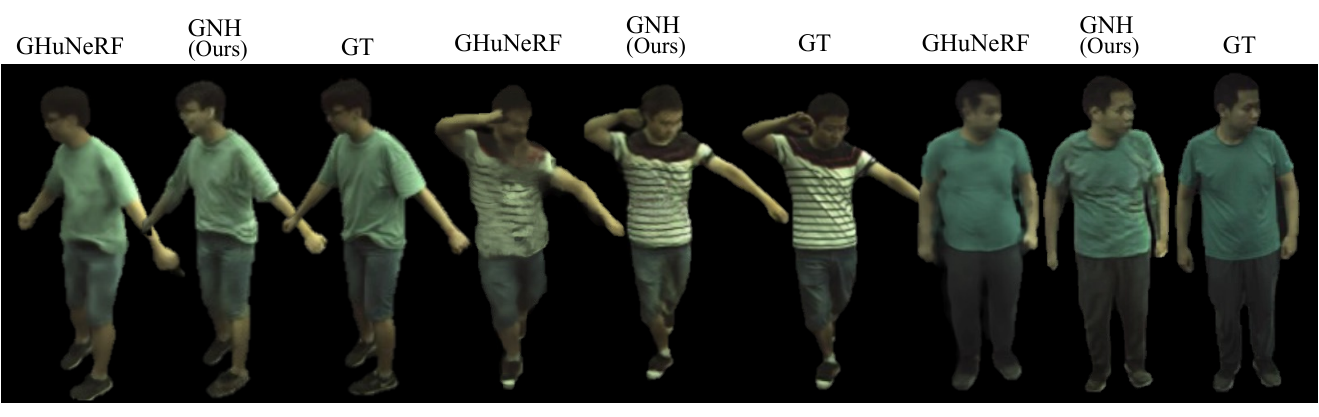}
    \captionsetup{aboveskip=0pt}\captionsetup{belowskip=0pt}
    \caption{
        Qualitative evaluation of \textbf{seen} subject in \textbf{unseen} pose result on the ZJU-MoCap dataset.
    }
    \label{fig:seen_subject}
\end{figure*}
\begin{table}[tb]
    \centering
    \captionsetup{aboveskip=0pt}\captionsetup{belowskip=0pt}
    \caption{Quantitative evaluation of the rendering result of \textbf{seen} subject (the six subject in train set) in \textbf{unseen} pose on the ZJU-MoCap dataset.}
    \scalebox{0.8}{
    \begin{tabular}{lccccc}
    \Xhline{3\arrayrulewidth}
         Model & PSNR$\uparrow$ & SSIM$\uparrow$ & LPIPS$\downarrow$ & Avg. $\downarrow$  \\ \hline 
         GHuNeRF \cite{li2023ghunerf} & 29.42 & 0.969 & 51.07 & 21.99 \\ 
         GNH (Ours)& \textbf{29.62} & \textbf{0.969} & \textbf{33.28} & \textbf{18.68}  \\ \hline

    \end{tabular}}
    \label{tab:seen_subject}
\end{table}

\begin{table}[tb]
    \centering
    \captionsetup{aboveskip=0pt}\captionsetup{belowskip=0pt}
    \caption{Ablations and comparisons. Our final model is underlined.}
    \scalebox{0.87}{
    \begin{tabular}{llccccc}
    \Xhline{3\arrayrulewidth}
                    \multicolumn{1}{c}{Experiment}              & \multicolumn{1}{c}{Method}  & PSNR$\uparrow$ & SSIM$\uparrow$ & LPIPS$\downarrow$ & \multicolumn{2}{c}{Avg. $\downarrow$}  \\ \hline \hline
        \multirow{5}{*}{Objective Function} & w/o $\mathcal{L}_\text{color}$ & 27.14 & 0.963  & 41.08 & 24.78 & (+0.83) \\
                                    & w/o $\mathcal{L}_\text{LPIPS}$ & 26.45 & 0.956  & 56.71 & 29.94 & (+5.99) \\
                                    & w/o $\mathcal{L}_\text{adv}$ & 27.35 & 0.965 & 41.70 & 24.30 & (+0.35) \\
                                    & w/o $ \mathcal{L}_\text{ab}$ & 27.24 & 0.963 & 42.00 & 24.81  &  (+0.86) \\
                                    & \ul{Full Model} & 27.58 & 0.963 & 41.26 & 23.95  & \\\hline 
        \multirow{3}{*}{Feature Encoder}& $\mathbf{F}_\text{fine}$ & 27.32 & 0.961 &  41.91 & 24.56 & (+0.61) \\
                                     & $\mathbf{F}_\text{coarse}$& 24.84 & 0.954 & 51.79 & 33.25 &  (+9.30)\\ 
                                     & \ul{$\mathbf{F}_\text{coarse}\oplus \mathbf{F}_\text{fine}$} & 27.58 & 0.963 & 41.26 & 23.95  & \\\hline
        \multirow{4}{*}{Metadata}   & w/o camera diff. & 27.16 & 0.962 & 42.38 & 25.09 & (+1.14) \\
                                    & w/o pose diff. & 27.51 & 0.961 & 42.35 & 24.52 & (+0.57) \\
                                    & w/o target pose & 27.16 & 0.963 & 42.11 & 24.97 &  (+1.02) \\
                                    & \ul{Full Model} & 27.58 & 0.963 & 41.26 & 23.95 &\\\hline
        \multirow{2}{*}{Normalization} & Batch Norm & 27.10 & 0.960 & 53.67 & 27.56 & (+ 3.61) \\
                                    & \ul{Group Norm} & 27.58 & 0.963 & 41.26 & 23.95 & \\ \hline
        \multirow{2}{*}{Rendering Module} & ResUNet\cite{zhang2018road} & 27.36 & 0.964 & 41.81 & 24.42 & (+0.47) \\                    
                                         & \ul{ResUNet++}\cite{jha2019resunetplus} & 27.58 & 0.963 & 41.26 & 23.95 & \\
        \hline
    \end{tabular}}
    \label{tab:ablation}
\end{table}

\textbf{Rendering Result of Seen Identities.}
We additionally present an evaluation of animatable rendering for \textit{seen} identities. In this experiment, we utilized videos from six subjects (313, 315, 377, 386, 390, 392) within our training set and assessed frames from each video's final part, which were not used in our generalizable training. The quantitative and qualitative results are shown in Table \ref{tab:seen_subject} and Fig. \ref{fig:seen_subject} respectively. For both metrics, our GNH surpassed the baseline method, showcasing superior performance.

\begin{figure*}[tb]
    \centering
    \includegraphics[width=0.65\linewidth]{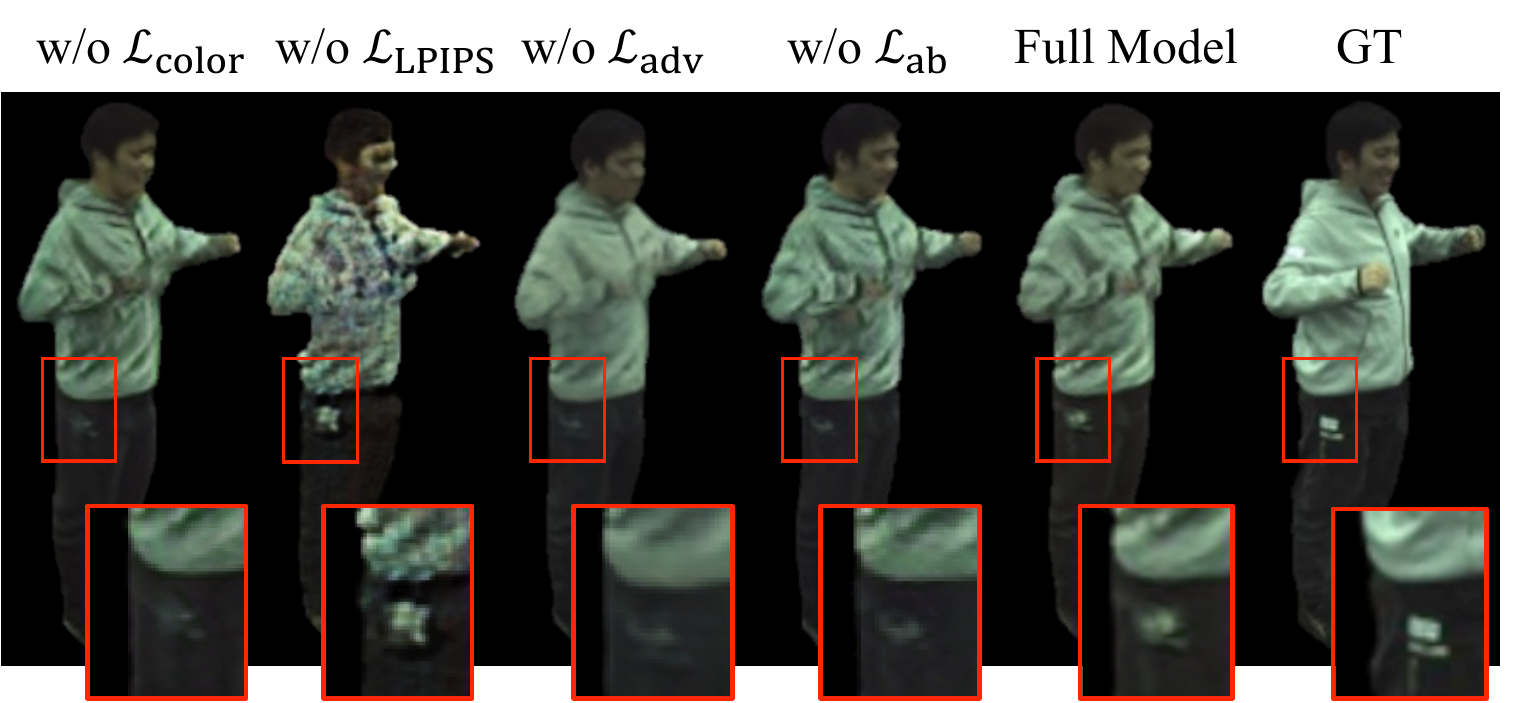}
    \captionsetup{aboveskip=3pt}\captionsetup{belowskip=0pt}
    \caption{
        Ablation study of the objective function.
    }
    \label{fig:loss_ablation}
\end{figure*}

\begin{figure}[tb]
\begin{minipage}[tb]{0.46\linewidth}
\centering
\includegraphics[width=1.0\linewidth]{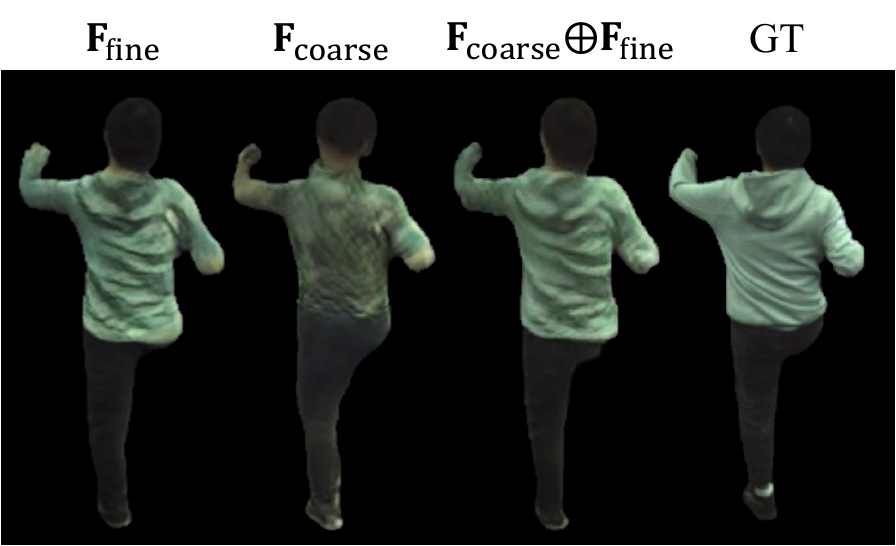}
\captionsetup{aboveskip=3pt}\captionsetup{belowskip=0pt}
\caption{Ablation study of the feature encoder module.}
\label{fig:feature_ablation}
\end{minipage}
\begin{minipage}[tb]{0.53\linewidth}
\centering
\includegraphics[width=1.0\linewidth]{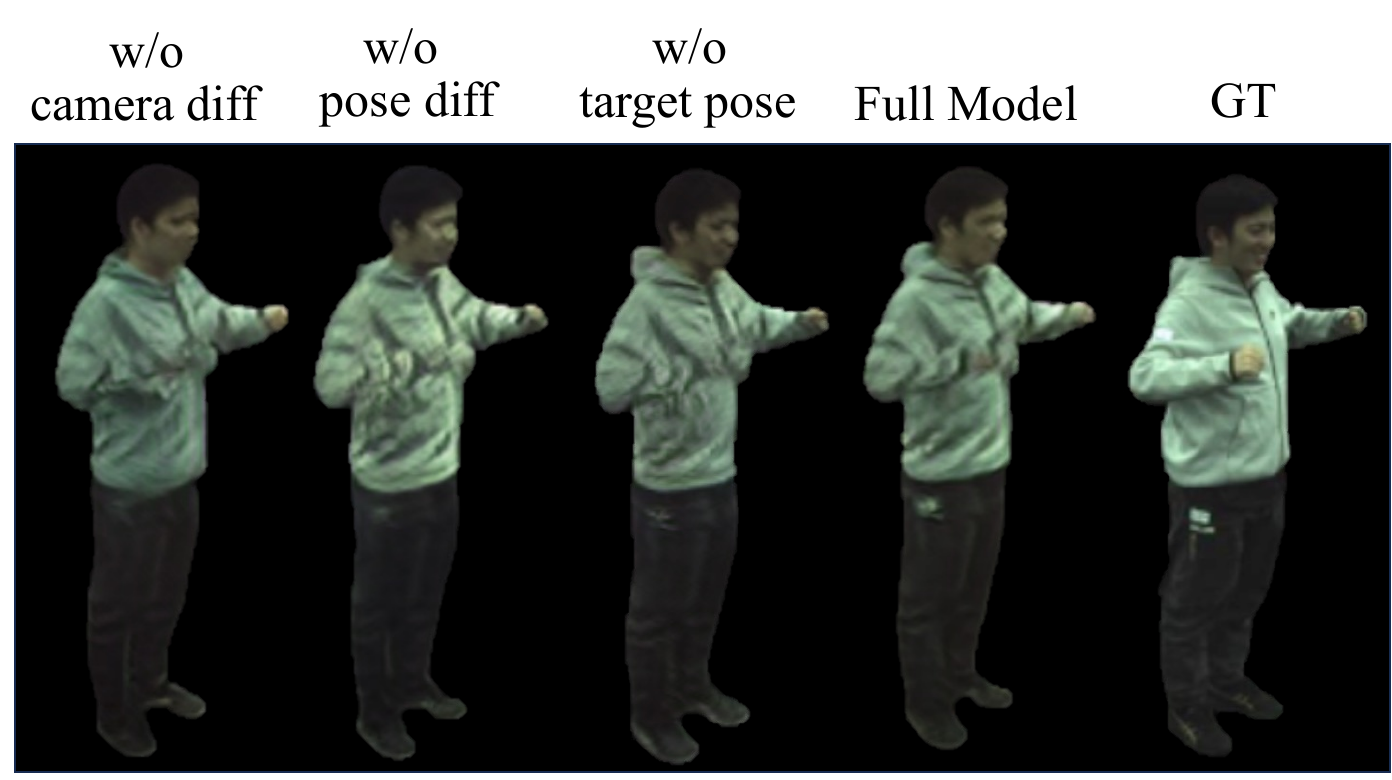}
\captionsetup{aboveskip=3pt}\captionsetup{belowskip=0pt}
\caption{Ablation study of the metadata for multi-frame fusion transformer.}
\label{fig:metadata_ablation}
\end{minipage}
\end{figure}

\subsection{Ablations and Other Analysis}
We conduct a series of ablation studies on the ZJU-MoCap dataset to validate the effectiveness of our architectural designs. The scores reported reflect the 5-shot results for subject 387 from the ZJU-MoCap dataset. The results are in Table \ref{tab:ablation}.

\textbf{Objective Function.}
To validate the effectiveness of our proposed optimization function, we demonstrate the effect of each element. In addition to the quantitative evaluation demonstrating that each element contributes to performance improvement, we also report the qualitative comparison in Fig. \ref{fig:loss_ablation}. The photometric loss $\mathcal{L}_\text{L1}$, the discriminative loss $\mathcal{L}_\text{Disc}$, and the anti-bias loss $\mathcal{L}_\text{ab}$ are contributing to the representation of detailed parts and the perspective loss $\mathcal{L}_\text{VGG}$ is contributing to the overall visual quality.

\textbf{Feature Encoder Design.}
To validate the effectiveness of our feature encoder which extracts appearance features from source frames, we demonstrate the effect of each element by removing elements. In addition to the quantitative evaluation showing that both course image feature $\mathbf{F}_\text{coarse}$ and fine image feature $\mathbf{F}_\text{fine}$ contribute to the performance improvement, we also report the qualitative comparison in Fig. \ref{fig:feature_ablation}.

\textbf{Metadata for the Multi-Frame Fusion Transformer.}
We ablate the metadata input for the multi-frame fusion transformer by demonstrating the effect of each element. In addition to the quantitative evaluation showing that all of the metadata contribute to improving the performance, we also report the qualitative comparison in Fig. \ref{fig:metadata_ablation}.


\textbf{Normalization Layer.}
Since we utilize the group normalization\cite{wu2018group} instead of the batch normalization used in the official implementation of ResUNet++\cite{jha2019resunetplus}, we compare the performance of these normalization layers. The quantitative evaluation shows that the group normalization improves the rendering performance by $23.1\%$ in LPIPS in our experiment.

\textbf{Rendering Module.}
We implement a variant of the image rendering module with ResUNet \cite{zhang2018road}. The result shows the overall performance of this version is lower than our final model.
\section{Discussion and Conclusion}
We introduce the \textit{Generalizable Neural Human Renderer} (GNH), a pioneering approach for rendering high-fidelity, animatable humans from monocular video inputs, eliminating the need for test-time optimization. GNH leverages a streamlined three-stage rendering process that not only simplifies the workflow but also significantly enhances the quality of the output. This method has been rigorously tested and has demonstrated superior performance, both qualitatively and quantitatively, across three dynamic human motion datasets, underscoring its robustness and versatility in handling complex animations.

\textbf{Limitations and Future Work.}
GNH skillfully utilizes source views to render accurate target views and poses, yet it relies on precise pose and mask estimations in input views. Although our approach is proficient in handling SMPL-based human topology transformations, it does not account for dynamic lighting changes, which could compromise its performance under varied lighting conditions. Acknowledging these limitations, we see substantial opportunities to improve pose and mask estimation accuracy and to enhance adaptability to diverse lighting conditions. These areas offer fertile ground for future research, aiming to enhance the robustness and applicability of GNH and similar methodologies.

\clearpage  

%
%
\bibliographystyle{splncs04}
\bibliography{egbib}

\newpage
\clearpage
\setcounter{section}{0}
\renewcommand{\thesection}{\Alph{section}}
\setcounter{figure}{0}
\renewcommand{\thefigure}{A\arabic{figure}}
\setcounter{table}{0}
\renewcommand{\thetable}{A\arabic{table}}
\setcounter{footnote}{0} 
\setcounter{equation}{0}

\section{Dataset Details}
This section includes the detailed datasets training setup.

\textbf{ZJU-MoCap Dataset.}
For the ZJU-MoCap dataset, we utilized the first 550 frames for the training.
For the unseen subject evaluation, we used the first 300 frames as the source input video and evaluated our model on the next 250 frames for every 10 frames.
For seen subject evaluation, we utilized the first 550 frames which is used for training as the source input video, and evaluated our model on the rest of the video for every 10 frames.

\textbf{People Snapshot Dataset.}
For the People Snapshot dataset, we utilized every five frames of the whole video for the training.
For the evaluation, we used every five frames from the first 300 frames for the source input video and evaluated our model on every 10 frames of the rest of the video.

\textbf{AIST++ Dataset.}
For the AIST++ dataset, we utilized every four frames from 200-800 frames of each video for the training.
For the evaluation, we used every four frames from 200-1400 frames for the source input video and evaluated our model on every four frames of the rest of the video.


\setcounter{figure}{0}
\renewcommand{\thefigure}{B\arabic{figure}}
\begin{figure*}[b]
    \centering
    \includegraphics[width=0.65\linewidth]{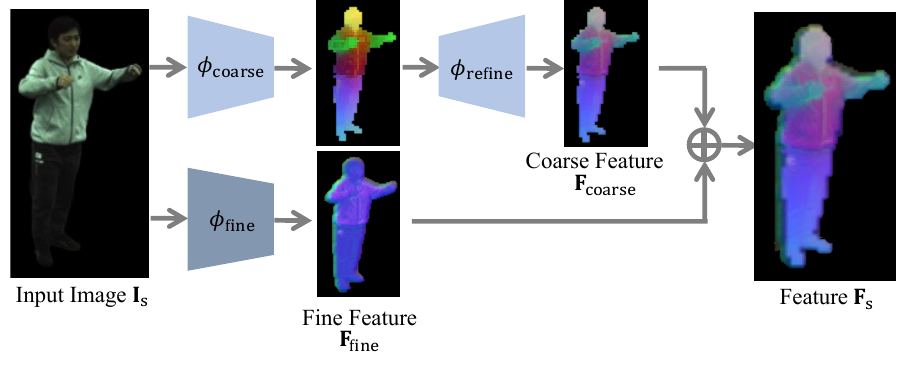}
    \captionsetup{aboveskip=0pt}\captionsetup{belowskip=0pt}
    \caption{
        Structure of our feature encoder module.
    }
    \label{fig:feature_encoder}
\end{figure*}
\section{Network Structure}
The detailed structure of our feature encoder module is shown in Fig. \ref{fig:feature_encoder}.

\hspace{\parindent}\textbf{Coarse Feature Refinement Net $\phi_\text{refine}$.}
We process the output coarse feature $\mathbf{F}_\text{coarse}$ with a convolutional transformer architecture.
We use patch embedding and four layers of efficient self-attention blocks from Segformer \cite{xie2021segformer}.
Three convolutional layers follow this transformer structure and produce final coarse feature maps $\mathbf{F}_\text{coarse}$ with dimensions of $256 \times 256 \times 96$.

\textbf{Fine Feature Encoder $\phi_\text{fine}$.}
In parallel to the coarse feature extraction process, we extract fine-grained features using a five-layer convolutional network, which includes only one downsampling layer. We use the Leaky Rectified Linear Unit (LeakyReLU) for activation. Following the convolutional network, the feature image is upsampled using bilinear interpolation, resulting in refined feature maps $\mathbf{F}_\text{fine}$ with dimensions of $256 \times 256 \times 96$.

\textbf{Multi-Frame Fusion Transformer $\Psi_\text{multi\_frame}$.}
To fuse feature maps from each source image into a single feature map, we use a transformer network equipped with cross-view attention \cite{varma2023gnt}. Considering that not all pixels possess distinct features, our approach carefully targets only those pixels with projected features, enhancing computational efficiency. The architecture of our transformer blocks is uniformly designed with a dimension of 256 and we set the transformer depth as four.



\end{document}